%% file: Limmer_ITSC2016.tex
\documentclass[letterpaper, 10 pt, conference]{ieeeconf}  

\IEEEoverridecommandlockouts                              

\usepackage{cite}
\usepackage{graphicx}
\usepackage{amsmath,amssymb} 
\usepackage{color}
\usepackage{pgfplots}
\usepackage{subfigure}
\usepackage[utf8]{inputenc}
\usepackage[T1]{fontenc}
\usepackage{lmodern}
\usepackage{stfloats}
\usepackage{booktabs}
\usepackage{url}
\usepackage{caption}

\usetikzlibrary{external}

\input{Macros}

\begin{document}

\title{\huge Robust Deep-Learning-Based Road-Prediction {\LARGE \\ for Augmented Reality Navigation Systems}}

\author{
Matthias Limmer$^{1}$*, 
Julian Forster$^{1}$*, 
Dennis Baudach$^{1}$, 
Florian Schüle$^{2}$, \\
Roland Schweiger$^{1}$ and
Hendrik P.A. Lensch$^{3}$%
\thanks{This work was partially funded by the European Commission under the ECSEL Joint Undertaking in the scope of the DESERVE project. {\tt http://www.deserve-project.eu/}}%
\thanks{$^{1}$ M. Limmer, J. Forster, D. Baudach and R. Schweiger are with Daimler AG R\&D, Ulm, Germany}%
\thanks{$^{2}$ F. Schüle is with the Institute of Measurement, Control and Microtechnology, University of Ulm, Germany}%
\thanks{$^{3}$ H. Lensch is with the Department of Computer Graphics, Eberhard Karls Universität, Tübingen, Germany}%
\thanks{*These authors contributed equally to this work}%
}

\maketitle
\thispagestyle{empty}
\pagestyle{empty}

\input{chapters/abstract}

\input{chapters/introduction}
\input{chapters/relwork}
\input{chapters/framework}
\input{chapters/roaddetect}
\input{chapters/experiments}
\input{chapters/summary}

\addtolength{\textheight}{-15cm}

\input{chapters/acknowledgments}

\bibliographystyle{IEEEtran}
\bibliography{bib}

\end{document}

%% file: Macros.tex
\newcommand{\relu}{\mathrm{ReLU}}

\newcommand{\kp}{k_p}
\newcommand{\kc}{k_c}
\newcommand{\nc}{n_c}
\newcommand{\nb}{n_b}
\newcommand{\nf}{n_f}
\newcommand{\np}{n_p}
\newcommand{\nl}{n_l}
\newcommand{\stride}{\mathrm{stride}}
\newcommand{\strideproduct}{\prod{\stride_{\kp}}}

\newcommand{\nallimages}{7095 }

\newcommand{\nkilometer}{13.5\,km }
\newcommand{\ntrainimages}{6895 }
\newcommand{\nevalimages}{200 }

\newcommand{\road}{\texttt{road}}
\newcommand{\vehicle}{\texttt{vehicle}}
\newcommand{\background}{\texttt{background}}
\newcommand{\sky}{\texttt{sky}}
\newcommand{\vru}{\texttt{vru}}
\newcommand{\infrastructure}{\texttt{infrastructure}}

\newcommand{\topo}[3]{\texttt{topo-#1-#2-#3}}

%% file: chapters/abstract.tex
\begin{abstract}
This paper proposes an approach that predicts the road course from camera sensors leveraging deep learning techniques.
Road pixels are identified by training a multi-scale convolutional neural network on a large number of full-scene-labeled night-time road images including adverse weather conditions.
A framework is presented that applies the proposed approach to longer distance road course estimation, which is the basis for an augmented reality navigation application.
In this framework long range sensor data (radar) and data from a map database are fused with short range sensor data (camera) to produce a precise longitudinal and lateral localization and road course estimation.
The proposed approach reliably detects roads with and without lane markings and thus increases the robustness and availability of road course estimations and augmented reality navigation.
Evaluations on an extensive set of high precision ground truth data taken from a differential GPS and an inertial measurement unit show that the proposed approach reaches state-of-the-art performance without the limitation of requiring existing lane markings.
\end{abstract}

%% file: chapters/introduction.tex
\section{Introduction}

Augmented reality navigation applications that support drivers navigating in unknown environments are one example of future \emph{advanced driver assistance systems} (ADAS).
Although this ADAS function is aimed mostly at urban navigation, where the difficulty lies in navigating in a complex road network, another use case is inter-urban navigation, especially for poor visibility conditions (e.g., fog, snow, night, \dots).
Regular navigation applications leverage a map database and a GPS sensor for coarse localization.
Augmented reality applications, however, not only require a precise localization but also a precise road course estimation.
Accurate lateral localization and shorter distance road course estimation is particularly important for realistic augmentations of the camera image.
Common approaches exploit existing lane and road markings for this task (cf.~\cite{Schuele2013,Deusch2014}).
Lane and road markings, though, might not be usable or available for all inter-urban roads because of damage, soiling or simple absence.

The image-based road detection approach presented in this paper classifies each pixel with a deep multi-scale \emph{convolutional neural network} (CNN).
The CNN learns feature extractors to identify road pixels in an integrated fashion.
It is therefore capable of reliably classifying road pixels disregarding the presence of lane markings.
Classified road pixels are homogenized by a floodfill algorithm to create a coherent road segment.
A road contour is extracted from that segment and fitted into a spline-based road model, the \emph{optical map}.
This optical map is fused with processed data from a map database, the \emph{digital map}, and a \emph{grid map} from a radar sensor to conduct a precise localization and road course estimation.
This is used in the augmented reality navigation system depicted in Fig.~\ref{fig:eyecatcher}.
\input{figures/eyecatcher}

The approach is trained on a large number of full-scene-labeled \emph{near infrared} (NIR) images showing night-time road scenes including adverse weather conditions.
Localization and road prediction results are evaluated in extensive experiments against ground truth trajectories measured by a high precision \emph{inertial measurement unit} (IMU) and \emph{differential GPS} (D-GPS). 
Evaluation results show state-of-the-art performance compared to a baseline approach~\cite{Risack1998}, but no failures when lane markings are not available.
This increases the robustness and availability of the application.

%% file: figures/eyecatcher.tex
\begin{figure}
\captionsetup{font=scriptsize}
\centering
\includegraphics[width=0.48\textwidth]{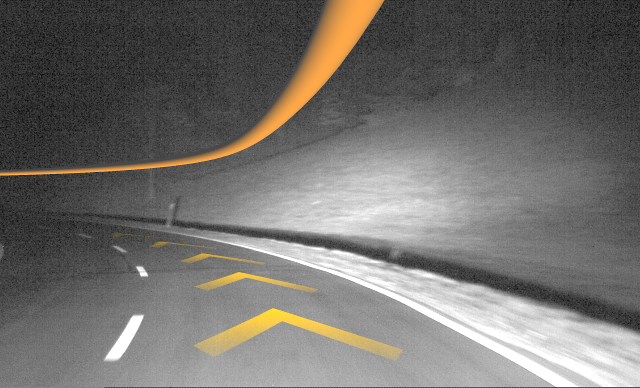}%
\caption{
An augmented reality navigation application. 
The orange tube displays the longer distance road course while the arrows are pinned to the road surface and augment the short distance road course.
Figures are best viewed in color.
}\label{fig:eyecatcher}
\vspace{-0.5\baselineskip}%
\end{figure}

%% file: chapters/relwork.tex
\section{Related Work}\label{sec:relwork}
Using digital maps as a source for road course estimation at longer distances requires accurate localization.
The precision of common GPS sensors of up to 10\,m for localization satisfies the needs of regular navigation systems but not those of a precise road course estimation, especially for augmented reality navigation~\cite{Schuele2013}. 
To achieve a higher precision for longitudinal and lateral localization, road course estimation applications fuse multiple sensors with longer and shorter perception ranges. 
An exhaustive overview of different sensor fusion approaches is collated in~\cite{BarHillel2012}.
In the following, a few approaches are introduced that are closer related to the scope of this paper.

Tsogas et al.~\cite{Tsogas2011} fuse measurements of a camera, laser scanner and a digital map based on the \emph{clothoid} road model. 
A Sugeno-fuzzy system determines appropriate weightings for each of the different sensors dependent on the prediction distance from the ego-vehicle and the range of the sensor.
The clothoid model, though, is only able to model cubic road curvatures.
Complex curvatures, which commonly reside in arbitrary rural roads, can only be represented by joining several clothoids together.
This, however, would increase the parameter space considerably and is not modeled by the aforementioned approach.

The sensor fusion system of Deusch et al.~\cite{Deusch2014} is not dependent on the clothoid model and the digital map.
It belongs to the category of systems that record a custom map containing landmarks and sensor data that can be used to localize the car later on.
Coordinates from a D-GPS sensor are mapped to landmarks extracted from forward and backward looking cameras and the occupancy grid of a laser scanner.
In a recall phase, the regular GPS-position is refined by matching concurrently extracted landmarks to those in the database. 
The creation of a landmark database, though, is a procedure that needs to be completed in advance.
Moreover, maintenance of the database has to be performed on a regular basis to remove landmark errors because of construction works, etc. 

Schüle et. al.~\cite{Schuele2013} describe a framework that fuses a NIR camera sensor, a radar sensor and a digital map.
It performs longitudinal localization by fusing a radar grid map and a digital map using a particle filter.
Precise lateral localization is then accomplished by fusing the longitudinally mapped digital map with an optical lane recognition algorithm~\cite{Risack1998} in the camera image.
In subsequent works~\cite{Schuele2013a}, a Bayesian fusion system that performs the final road course estimation is introduced.
In both systems, the road course model is not a clothoid but rather lists of connected 2D points sampling the right and left borders of the lane.
This approach, as well as all aforementioned approaches, relies on lane marking detectors for the estimation of an optical map.
To increase the robustness and availability of such a system, an optical road course recognition is desired that works independent of lane or road markings.

Seo et al.~\cite{Seo2014} describe a road boundary estimator based on intensity distribution thresholding from camera images.
The thresholded intensity distribution is extracted from a \emph{region of interest} (ROI) on the inverse perspective mapped camera image.
Extracted road boundaries are tracked over time by a Bayes filter.
Although the thresholding method is a simple and efficient approach for detecting road pixels, it might fail for roads with a high illumination variance (e.g. containing sharp shadows).  

Fernández et al.~\cite{Fernandez2015} perform road detection by training decision trees.
They use the disparity features of a stereo camera for a ground plane detection and several hand-crafted color and texture features to classify superpixels segmented by a watershed transform.
This approach, though, strongly relies on features not available for grayscale monocular camera images.

Alvarez et al.~\cite{Alvarez2012} describe a road scene segmentation from single images using a convolutional neural network.
The CNN is trained on publicly available annotated road scenes that are not necessarily images from the camera used in the application.
To overcome this and allow adapting to immediate situations, the CNN classification is fused with the color intensity distribution from an ROI ahead of the vehicle through a Bayesian framework.
Recent developments of CNN classifiers, though, show that such a fusion step is not necessary if a network is pre-trained on a large amount of data from a similar sensor and only adjusted to the current sensor by providing a smaller set of training data (c.v.~\cite{Long2015,Cordts2016}).
Apart from that, this approach is not directly suitable for a road course extraction.
Other road users that possibly occlude parts of the road are not explicitly classified, what complicates road border extractions in these cases.

The framework proposed in this paper is based on \cite{Schuele2013}, but replaces the optical lane detection module from~\cite{Risack1998} with a road segmentation module based on deep multi-scale CNNs.
It is trained on a dataset of night-time images with a large variety of road and weather situations with and without lane markings.
This approach therefore increases the robustness and availability of shorter distance road course estimations to situations without lane markings or adverse weather situations.

The remainder of this paper is structured as follows: 
Section~\ref{sec:framework} describes the framework, while Section~\ref{sec:roaddetect} presents the road detection module in more detail. 
Extensive experiments are described and discussed in Section~\ref{sec:experiments}.
Section~\ref{sec:conclusion} summarizes the results.

%% file: chapters/framework.tex
\section{Framework}\label{sec:framework}
To compute a reliable localization and road course estimation, a framework that fuses different sensor inputs is needed.
This paper leverages a derivation of the framework from~\cite{Schuele2013} and is depicted in Fig.~\ref{fig:framework}.
The modules of the framework are as follows:
First, radar data in combination with a tracked ego-motion estimation produces a grid map.
Second, initialized by the GPS position, the rough location in a commercially available map database is determined and map parameters for that location are transformed into a compliant digital map model.
Third, the grid map and the digital map are fused to produce a longitudinally matched digital map.
The fourth module performs road detection in a corresponding camera image and produces an optical map.
\input{figures/framework}

\subsection{Grid Mapping}\label{sec:gridmap}
A grid map is a 2D map representing the local environment quantized into equally sized cells representing occupancy (see Fig.~\ref{fig:gridmap}).
Each cell temporally integrates respective sensor measurements from a distance measuring sensor and thereby reduces the inherent noise and uncertainties of singular measurements.
In the proposed framework, data from an imaging automotive radar, which returns both, the distances of reflections and their velocities, is stored in the grid map.
Since the ego-vehicle is moving, its relative position on the grid map needs to be determined by estimating the ego-motion.
An extended Kalman filter with a CTRV-model (constant turn rate and velocity~\cite{Schubert2008}) leverages the wheel speeds and yaw rate measurements to accomplish an ego-motion estimation.
The ego-motion estimation is then used to determine the correct cells where static radar objects are stored and integrated over time.

\subsection{Digital Mapping}\label{sec:digmap}
A commercial map database commonly stores its information in annotated discrete shape points using the UTM (Universal Transverse Mercator) coordinate system.
The amount of points per road, the accuracy of such points and the meta-information per point varies greatly, since major roads are better sampled and maintained by database providers.
To obtain a continuous local digital road model, shape points around the current ego-vehicle's location are interpolated by a cubic hermite spline.
This creates the digital map, which serves as the base for the following fusion modules.

\subsection{Map Matching}\label{sec:mapmatch}
To estimate the orientation and longitudinal position of the ego-vehicle on the digital map, the grid map is fitted into the digital map using a particle filter.
Each particle of the filter represents the position and orientation of the vehicle and is weighted by how well the digital map and the grid map fit using various features~\cite{Szczot2010}.
The sampling of the particles is initialized by the previous position or the GPS position if no previous position is available.

\subsection{Road Detection}\label{sec:lanedetect}
The original \emph{Optical Lane Detection} module~\cite{Risack1998} in the framework of~\cite{Schuele2013} is replaced with the lane-independent \emph{Road Detection} module proposed in this paper.
In this processing step, pixels in a camera image belonging to the currently traveled road are identified.
These detected pixels are used to determine the road boundaries which are then transformed into and tracked by the optical map. 
Further details of this processing step are described in Section~\ref{sec:roaddetect}.

\subsection{Lane Course Fusion}\label{sec:lanefuse}
To increase the precision of the lateral localization, the optical map is fused with the digital map.
Therefore, lateral coordinates in the ego-vehicle's coordinate system are sampled from both maps along the longitudinal trajectories of lane or road borders. 
Corresponding lateral positions are linearly interpolated by weighting each sensor according to its reliability for different distances from the ego vehicle.
The optical map is very reliable for close distances while the digital map is more reliable for larger distances.
The specific weighting scheme is described in~\cite{Schuele2013}.
\input{figures/gridmap}

%% file: figures/framework.tex
\begin{figure}
\captionsetup{font=scriptsize}
\centering
\includegraphics[width=0.48\textwidth]{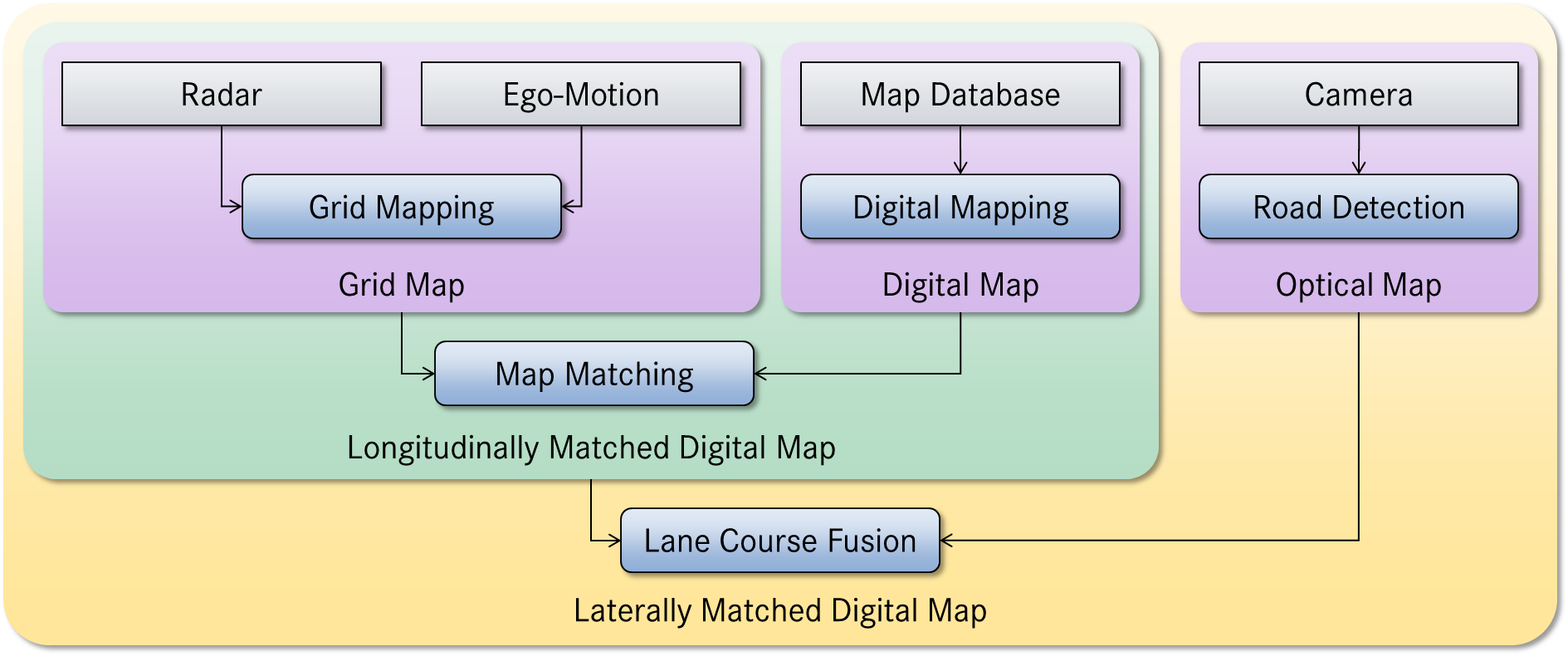}%
\caption{Processing units of the proposed framework.}\label{fig:framework}
\end{figure}

%% file: figures/gridmap.tex
\begin{figure}
\captionsetup{font=scriptsize}
\centering
\includegraphics[width=0.48\textwidth]{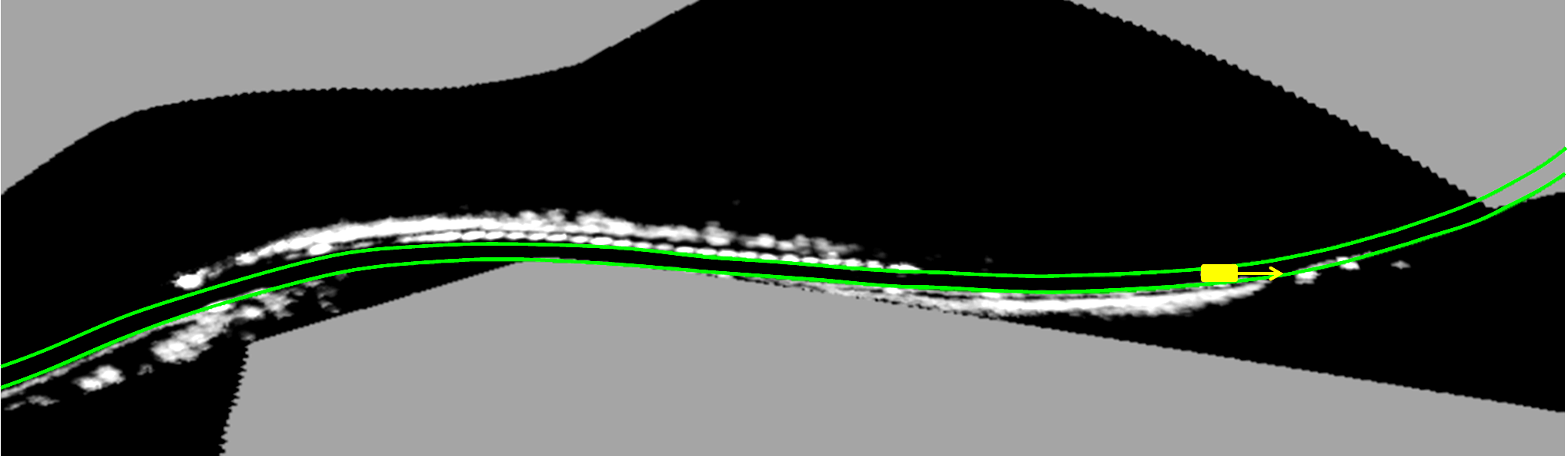}%
\caption{
An example of a grid map. 
The ego-vehicle and its travel direction are displayed as the orange box with the arrow. 
The white and gray dots show integrated values of reflections from the radar sensor. 
These reflections are generated primarily by grass, curbs or barriers at the side of the road. 
The green lines are splines of the digital map road shape points matched to the grid map.
}\label{fig:gridmap}
\end{figure}

%% file: chapters/roaddetect.tex
\section{Road Detection}\label{sec:roaddetect}

The road detection module described in the following identifies the currently traveled road in a camera image by performing a pixel classification using deep learning techniques.
It then extracts and tracks the left and right road border taking into account uncertainties and border-occlusions by other road users.
It then computes the optical map that can be fused with the digital map.
\input{figures/netwarch}

\subsection{Scene Labeling}\label{sec:SL}
The scene labeling module proposed in this paper is a deep multi-scale CNN.
It combines the approach of~\cite{Simonyan2014} with the multi-scale scheme of~\cite{Farabet2013}.
\cite{Simonyan2014} introduces network topologies characterized by many convolution layers with small convolution kernels and comparatively few pooling layers.
Many convolution layers increase the amount of non-linearities and thus the capability of the network to learn complex classification functions. 
If small convolution kernel sizes are used, the increase of convolution layers does not necessarily lead to a drastic increase of computational complexity.
Multiple scales further improve the scale-invariance of the network without increasing the depth of the network.
Since real-time performance is needed for augmented reality applications, techniques from~\cite{Giusti2013} and~\cite{Thom2016} are implemented for a computational efficient application of a CNN to entire images.

Multi-scale neural networks process multiple scales of the same input data concurrently.
An image pyramid of $\nl$ levels is constructed by reducing the image resolution by $0.5$ in both dimensions for each new level.
Each pyramid level is normalized to zero-mean unit-variance in a local neighborhood, which enhances the texture and equalizes bright and dark areas in the image.
The normalized image pyramid levels are then fed to their respective branches of the neural network. 
All branches of the network are constructed with the same structure and are finally joined in a fully-connected layer that also serves as the output layer of the network.
A diagram of possible network topologies of the above defined multi-scale CNN is depicted in Fig.~\ref{fig:netwarch}.

Though each branch is structurally identical, no weights are shared between the branches. 
A branch overall consists of alternating $\np$ pooling layers and $\nb = \np+1$ \emph{convolution layer blocks}. 
Every convolution layer block consists of $\nc$ convolution layers that use the $\relu$ function: $\relu(x) = \max(0,x)$ as activation function.
The size of the filter bank $\nf$ is identical within each convolution layer block and is doubled after each pooling layer. 
The kernel size of the convolution kernels $\kc$ is the same in all convolution layers.

In a patch-based application of the proposed network, correctly sized image patches need to be extracted from the normalized image pyramid levels prior to feeding them to the branches.
Applying the CNN efficiently to complete images while retaining the full image resolution requires slight changes in various layers and the introduction of several helper layers into the network (see~\cite{Giusti2013,Thom2016}).
These changes are explained in the following.

\subsubsection{Overlapping Pooling}
In an image-based application, pooling layers, which are normally strided ($\strideproduct > 1$) according to their kernel size $\kp$, need to be applied in an overlapping fashion ($\strideproduct = 1$), so that no resolution is lost. 

\subsubsection{Fragmentation}
Fragmentation layers need to be inserted after each pooling layer. 
They split the oversized feature maps of the preceding layer into $\strideproduct$ feature maps of reduced resolution, which are processed individually afterwards. 
This ensures that the subsampling property of the pooling functions is preserved without loosing resolution.
Fig.~\ref{fig:fragmentation} depicts a $2\times 2$ fragmentation.
\input{figures/fragmentation}%

\subsubsection{Defragmentation}
A defragmentation layer is needed after the last convolution block before all branches are joined.
It reverts all performed fragmentations and transforms the fragmented feature map arrays into one cohesive feature map array that has the same%
\footnote{Valid convolutions might crop some border pixels.} 
resolution as the corresponding input pyramid level.

\subsubsection{Upscaling}
After defragmentation, the feature map arrays of lower pyramid levels need to be sampled up and eventually cropped so that they match the resolution of the lowest pyramid level.
In this manner, the feature maps of all scales can be concatenated and used as an input to the fully-connected layer.

\subsubsection{Convolutional Fully-Connected Layer} 
The fully-connected layer is applied in a convolutional fashion to emulate the patch-based functionality.
This transforms fully-connected layers into $1\times 1$ convolution layers with as many input channels as incoming feature maps.

\subsection{Road Segmentation}\label{sec:roadseg}
A CNN, such as outlined above, generates a class membership map, in which every pixel is assigned to one of the trained classes.
The class membership map of the preceding step needs to be segmented such that a cohesive road segment can be extracted.
Fig.~\ref{fig:roadseg} displays a road segmentation generated by the following steps.

Assuming that the biggest connected group of classified road pixels approximates the actual connected group of road pixels, detached road pixel clusters can be neglected.
Holes in the connected group of pixels are then filled leveraging a flood-fill algorithm.
The algorithm is seeded at the bottom of the image, since that is supposed to be part of the road in most of the cases.

A contour is extracted from the segmented road pixels by using the snake algorithm of~\cite{Pavlidis1982}.
The left and right road border is then determined by splitting the contour in half at the highest central contour point.
The road contours are ignored at all border pixels of the camera image.  
Contour pixels that are adjacent to pixels classified as other road-users, such as \vehicle\ pixels, are ignored as well, since road users might conceal parts of the correct road border.
Finally, the remaining contour pixels are transformed into the digital map's coordinate system and stored for tracking in the following frames.
\input{figures/roadseg}

\subsection{Road Border Shaping}\label{sec:bordershape}
To compute the optical map needed in the following fusion step, the tracked road border estimates need to be fitted into a conclusive road model.
All estimates are longitudinally binned, with each bin representing a road border shape point.
The values of each bin are analyzed to compute a reliability measure of that particular road border shape point.
The bin medians are used as the shape points for fitting a spline, while the interquartile range determines if a shape point is used in the spline computation. 
Exploiting meta-information contained in the digital map, other track splines, like lane borders, can be interpolated from the left and the right border splines.
If preceding processing steps continuously fail to deliver usable measurements for one road border (e.g., in sharp curves) that road border can be extrapolated by the other road border and the other track splines.
Fig.~\ref{fig:bordershape} displays a spline fitting for unreliable measurements.
\input{figures/bordershape}

%% file: figures/netwarch.tex
\begin{figure*}
\centering
\captionsetup{font=scriptsize}
\includegraphics[width=0.98\textwidth]{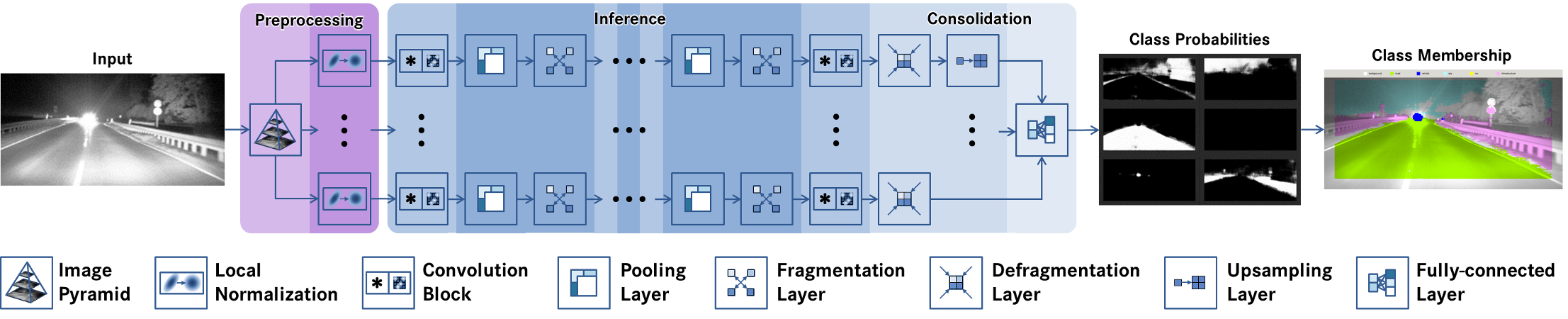}%
\caption{
Diagram of possible network architectures.
A preprocessing step generates a normalized image pyramid of $\nl$ levels.
Each level performs an inference in its own CNN branch that consists of alternating convolution blocks and pooling/fragmentation layers.
The fragmented feature maps of all branches are defragmented, upscaled and consolidated in a convolutional fully-connected layer that produces a \emph{probability map} for each class.
From these probability maps a final pixel classification is generated, the \emph{class membership map}.
}\label{fig:netwarch}
\vspace{-0.5\baselineskip}%
\end{figure*}

%% file: figures/fragmentation.tex
\begin{figure}
\captionsetup{font=scriptsize}
\centering
\includegraphics[width=0.48\textwidth]{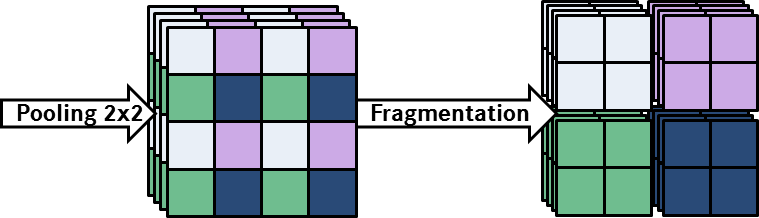}%
\caption{
Fragmentation of a $2\times 2$ pooling. 
The interleaved feature map pixels of an overlapping pooling are reordered to produce 4 independent subsampled feature map arrays.
}\label{fig:fragmentation}
\end{figure}

%% file: figures/roadseg.tex
\begin{figure}
\captionsetup{font=scriptsize}
\centering
\includegraphics[width=0.48\textwidth]{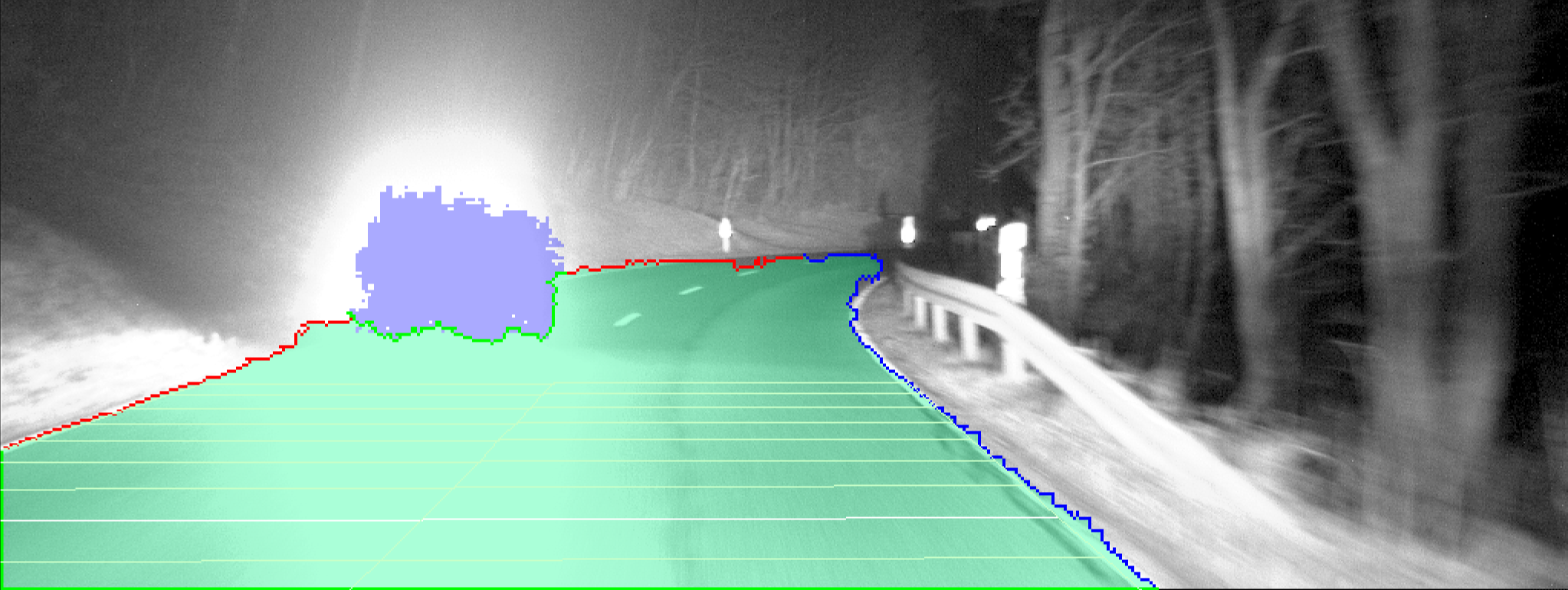}%
\caption{
The green area displays a road segmentation.
The blue area depicts a detected vehicle. 
The red and the blue lines denote the left and right border of the detected road.
The green lines (e.g. adjacent to the vehicle area) denote ignored border pixels.
}\label{fig:roadseg}
\end{figure}

%% file: figures/bordershape.tex
\begin{figure}
\captionsetup{font=scriptsize}
\centering
\includegraphics[width=0.48\textwidth]{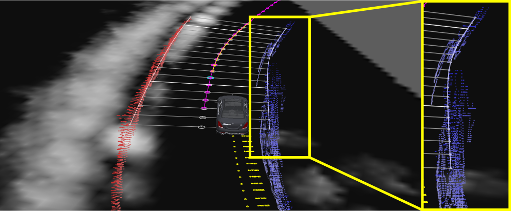}%
\caption{
Road border and center estimates projected to the grid map. 
The right border shows a big variance and missing values over time.
So the shape points of the right border are supported by the center points and the left border.
}\label{fig:bordershape}
\end{figure}

%% file: chapters/experiments.tex
\section{Experiments}\label{sec:experiments}
The performed experiments are twofold.
First, various network topologies were redundantly trained and evaluated with respect to their classification performance (Section~\ref{sec:SLresult}).
The dataset for training and evaluating the classifiers consists of \nallimages full scene labeled images from an NIR camera of rural road sequences at night containing a large variety of weather situations, seasons and landscapes.
Second, the optical map of the best performing classifiers were compared to the standard optical map from~\cite{Risack1998} using the fusion framework from~\cite{Schuele2013} (Section~\ref{sec:LCPresult}).
The evaluation is performed on five night-time sequences resulting in \nkilometer of driven distance with a ground-truth trajectory taken from a D-GPS sensor and a high accurate IMU.
Fig.~\ref{fig:seqs} shows examples images of these sequences.
\input{figures/seqs} 

\subsection{Training of the CNN}\label{sec:train}
Only certain combinations of parameters mentioned in section~\ref{sec:SL} are used in the experiments.
The influence of the number of pyramid levels ($\nl$), the deepness of the network while retaining the input patch size ($\nc, \kc$) and the initial number of filters of the first convolution block ($\nf$) were evaluated.
The topologies are therefore denoted as \topo{$\nl$}{$\nc$}{$\nf$}, with a parameter range of $\nl\in[1..5],\;(\nc,\kc)\in\{(1,7),(3,3)\}$ and $\nf\in\{16,32\}$.
Parameters $\nc$ and $\kc$ have to be chosen such that the input patch size stays the same, which holds for the above defined tuples.
The kernel size of the max pooling layers is fixed at $2\times 2$ pixels for all pooling layers in all topology variants.
Topology \topo{4}{1}{32}, for example, has the following parameters: $\nl=4,\;(\nc,\kc)=(1,7),\;\nf=32$.

All scene-labeled images are split into a set of \ntrainimages images for training and a set of \nevalimages images for evaluation.
To train the topologies, multinomial logistic regression performs a stochastic gradient descent leveraging the backpropagation algorithm~\cite{LeCun1998b} with linear learn rate annealing.
The target classes consist of the default class \background\ and specific classes: \road, \vehicle, \sky, \vru~(vulnerable road users) and \infrastructure.
Training examples are sampled patch-wise and class-balanced for an equal but random distribution of examples per class.
The trainable parameters of the networks are initialized by random-sampling a Gaussian distribution.
Learn rates for each topology are empirically determined by choosing the best performing learn rate in various mini-trainings.
With the selected learn rate full trainings are performed.
After completion, the biases of the fully connected layers are adjusted such that the multi-class extension of the \emph{Matthews Correlation Coefficient} (MCC)~\cite{Jurman2012} is optimized.
All trainings were conducted with \texttt{cuda-convnet}~\cite{Krizhevsky2012}. 

\subsection{Scene Labeling Results}\label{sec:SLresult}
Table~\ref{tab:SLresults} shows the classifier performances with regard to several measures.
These measures are the MCC~\cite{Jurman2012}, the overall accuracy (ACC), the intersection over union (IU) as an average over all classes ($\mathrm{IU_{global}}$) and specifically for the \road\ ($\mathrm{IU_{road}}$) and \vehicle\ class ($\mathrm{IU_{veh}}$).
The IU measure is defined as:
\begin{equation}
\mathrm{IU}=\frac{\mathrm{TP}}{\mathrm{TP}\cup\mathrm{FP}\cup\mathrm{FN}}
\end{equation}
where TP (true positives) is the amount of correctly classified pixels and $\mathrm{FP}\cup\mathrm{FN}$ (false positives and false negatives) the amount of wrongly classified pixels regarding one specific class.
To ensure that equal topologies perform similarly, each topology is trained three times.
The table shows the average result for one topology of the individually evaluated classifiers.
Topology \topo{1}{1}{16} is comparable to the best performing topology of~\cite{Alvarez2012}.
According to Table~\ref{tab:SLresults}, this topology achieves the lowest performance.
Other topologies are therefore encouraged for road segmentation tasks.

Topologies \topo{[3..5]}{3}{32} perform best regarding most of the measures.
This implies that an increase of pyramid levels after level 3 has almost no effect to the best performing topology variant.
Fig.~\ref{fig:mccgraphs} shows this effect in a graphical display of the MCC performances dependent on the pyramid levels.

\input{figures/SLresults}
\input{figures/MCCGraphs}

\subsection{Lane Course Prediction Results}\label{sec:LCPresult}
In the following, the fusion system performance is evaluated using the best performing classifier for each topology variant in relation to the optical lane recognition from~\cite{Risack1998} and the system without the optical map.
The performance measure is taken from~\cite{Schuele2013}.
It compares the deviations of the road course estimations from the ground truth trajectory for different distances to the ego-vehicle.
Since~\cite{Schuele2013} have shown that their fusion algorithm benefits primarily short range estimations, only the average performances of the five sequences for short range estimations (0-30\,m) are displayed in Table~\ref{tab:LCresults}.
The final row displays the failure rates of the lane-based recognition (percentage of frames, where no lane markings are detected).
It should be noted that the lane-based recognition measure is solely computed for frames, which contain detected lane markings.

Table~\ref{tab:LCresults} displays that the CNN-based optical map approach performs slightly worse for sequences containing good lane markings ($\mathcal{A,B}$), but better for sequences with bad weather ($\mathcal{C,D}$) or bad lane markings ($\mathcal{E}$).
Considering the better performing topologies (\topo{[3..5]}{*}{*}), the CNN-based optical maps show a similar range of performance values (\textasciitilde\,28\,cm) for sequences $\mathcal{A}$-$\mathcal{C}$ and $\mathcal{E}$.
Contrary to that, the lane-based approach shows a greater variance there, ranging from 19\,cm (seq. $\mathcal{A}$) to 50\,cm (seq. $\mathcal{C}$).
This implies that the CNN-based approach is performing more robustly than the lane-based approach, although the peak performance of the latter might not be reached. 

Sequence $\mathcal{D}$ is an exception regarding this robustness.
Its performance ranges from 50\,cm to 60\,cm for the CNN-based and 89\,cm for the lane-based approach.
This sequence contains severe snowfall and thus, poses a big challenge to sensor processing and detection algorithms.
While the CNN-based approach robustly delivers road course estimations for all frames (see Fig.~\ref{fig:eyecatcher2} for an example augmented image), the failure rate of the lane marking detection exceeds 90\,\%.
This means that the lane-based approach is practically inapplicable and needs to switch to a mode without optical map.
The performance of that mode, though, is always much lower than with an optical map (see first row of Table~\ref{tab:LCresults}).

Further, it should be noted that the best performing classifiers are not necessarily the best performing road course estimators, although a trend can be detected, when comparing topologies \topo{[1,2]}{*}{*} and \topo{[3..5]}{*}{*}.

\input{figures/LCresults}
\input{figures/eyecatcher2}

%% file: figures/seqs.tex
\begin{figure*}
\captionsetup{font=scriptsize}
\centering
\begin{tabular}{c@{}c@{}c@{}c@{}c}
{\scriptsize$\mathcal{A}$: regular} & {\scriptsize$\mathcal{B}$: wet road} & {\scriptsize$\mathcal{C}$: rain} & {\scriptsize$\mathcal{D}$: heavy snowfall} & {\scriptsize$\mathcal{E}$: bad lane markings} \\
\begin{minipage}{0.196\textwidth}
\includegraphics[width=\textwidth]{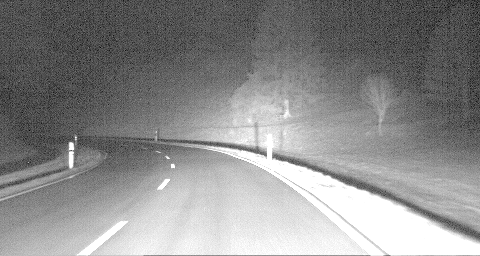}%
\end{minipage} &
\begin{minipage}{0.196\textwidth}
\includegraphics[width=\textwidth]{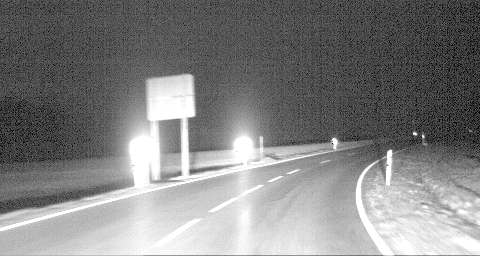}%
\end{minipage} &
\begin{minipage}{0.196\textwidth}
\includegraphics[width=\textwidth]{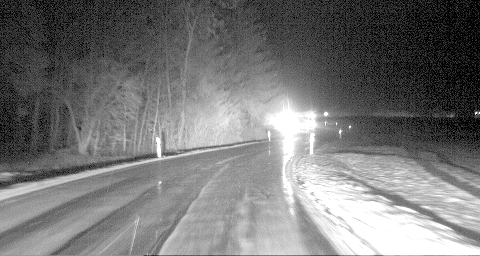}%
\end{minipage} &
\begin{minipage}{0.196\textwidth}
\includegraphics[width=\textwidth]{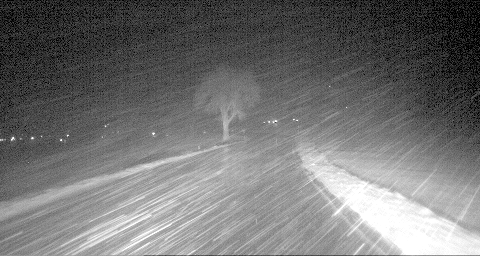}%
\end{minipage} &
\begin{minipage}{0.196\textwidth}
\includegraphics[width=\textwidth]{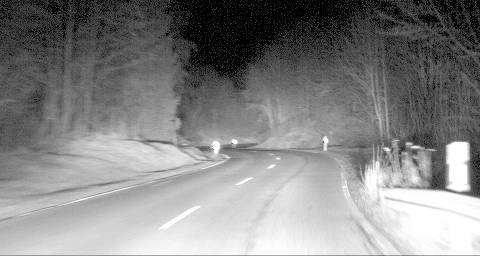}%
\end{minipage} \\
\end{tabular}\\
\caption{
Example images for the five evaluation sequences showing various road and weather conditions.
}\label{fig:seqs}
\end{figure*}

%% file: figures/SLresults.tex
\begin{table}
\captionsetup{font=scriptsize}
\centering
\caption{
Performance evaluation of the trained network topologies with respect to various measures.
Best performing values are marked in bold font.
}\label{tab:SLresults}
\begin{tabular}{c|ccccc}
  Name & MCC & ACC & $\mathrm{IU_{global}}$ & $\mathrm{IU_{road}}$ & $\mathrm{IU_{veh}}$ \\
  \midrule
  \topo{1}{1}{16}  & 0.56 & 0.69 & 0.40 & 0.67 & 0.31\\
  \topo{1}{1}{32}  & 0.60 & 0.71 & 0.43 & 0.70 & 0.38\\
  \topo{1}{3}{16}  & 0.61 & 0.72 & 0.44 & 0.70 & 0.38\\
  \topo{1}{3}{32}  & 0.66 & 0.75 & 0.48 & 0.75 & 0.45\\
  \topo{2}{1}{16}  & 0.66 & 0.76 & 0.48 & 0.78 & 0.42\\
  \topo{2}{1}{32}  & 0.70 & 0.78 & 0.51 & 0.82 & 0.48\\
  \topo{2}{3}{16}  & 0.71 & 0.79 & 0.52 & 0.82 & 0.49\\
  \topo{2}{3}{32}  & 0.72 & 0.80 & 0.54 & 0.83 & 0.52\\
  \topo{3}{1}{16}  & 0.71 & 0.79 & 0.52 & 0.84 & 0.48\\
  \topo{3}{1}{32}  & 0.73 & 0.81 & 0.54 & 0.86 & 0.52\\
  \topo{3}{3}{16}  & 0.75 & $\mathbf{0.82}$ & 0.56 & 0.86 & 0.55\\
  \topo{3}{3}{32}  & $\mathbf{0.76}$ & $\mathbf{0.83}$ & $\mathbf{0.58}$ & $\mathbf{0.87}$ & 0.57\\
  \topo{4}{1}{16}  & 0.73 & 0.80 & 0.53 & 0.86 & 0.46\\
  \topo{4}{1}{32}  & 0.73 & 0.81 & 0.53 & 0.86 & 0.49\\
  \topo{4}{3}{16}  & 0.75 & $\mathbf{0.82}$ & 0.56 & 0.86 & 0.56\\
  \topo{4}{3}{32}  & $\mathbf{0.77}$ & $\mathbf{0.83}$ & $\mathbf{0.59}$ & $\mathbf{0.88}$ & $\mathbf{0.58}$\\
  \topo{5}{1}{16}  & 0.71 & 0.79 & 0.50 & 0.85 & 0.43\\
  \topo{5}{1}{32}  & 0.71 & 0.79 & 0.51 & 0.86 & 0.46\\
  \topo{5}{3}{16}  & 0.75 & $\mathbf{0.82}$ & 0.56 & $\mathbf{0.87}$ & 0.56\\
  \topo{5}{3}{32}  & $\mathbf{0.77}$ & $\mathbf{0.83}$ & 0.57 & $\mathbf{0.88}$ & $\mathbf{0.59}$\\
\end{tabular}%
\end{table}

%% file: figures/MCCGraphs.tex
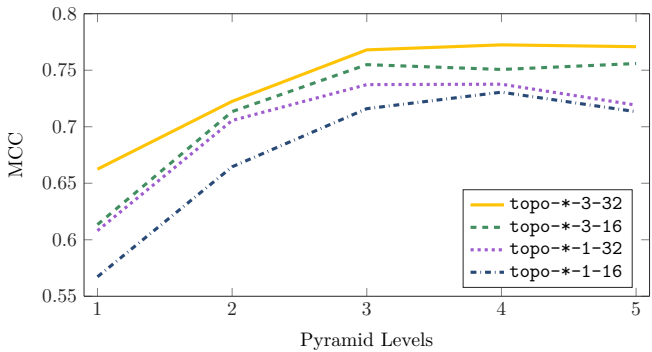
\begin{figure}
\captionsetup{font=scriptsize}
	\tikzsetnextfilename{MCCGraphs/MCCGraphs}
	\input{figures/MCCGraphs.tikz}
	\caption{
	The MCC of various topology variants in relation to the amount of pyramid levels.
	}
	\label{fig:mccgraphs}
\end{figure}

%% file: figures/MCCGraphs.tikz
%
%
%
\definecolor{myblue}{rgb}{0.16078431372549019607843137254902, 0.29019607843137254901960784313725, 0.46666666666666666666666666666667}%
\definecolor{mypurple}{rgb}{0.61960784313725490196078431372549, 0.36470588235294117647058823529412, 0.81176470588235294117647058823529}%
\definecolor{mygreen}{rgb}{0.25490196078431372549019607843137, 0.56078431372549019607843137254902, 0.38039215686274509803921568627451}%
\definecolor{myyellow}{rgb}{1.0,0.76470588235294117647058823529412,0.01960784313725490196078431372549}%
\begin{tikzpicture}[scale=0.74]

\begin{axis}[%
width=4in,
height=2in,
scale only axis,
separate axis lines,
every outer x axis line/.append style={white!15!black},
every x tick label/.append style={font=\color{white!15!black}},
xmin=0.9,
xmax=5.1,
xtick={1,2,3,4,5},
xlabel={Pyramid Levels},
every outer y axis line/.append style={white!15!black},
every y tick label/.append style={font=\color{white!15!black}},
ymin=0.55,
ymax=0.8,
ylabel={MCC},
legend style={draw=white!15!black,fill=white,legend cell align=left},
legend pos=south east
]
\addplot [color=myyellow,solid,style=ultra thick]
  table[row sep=crcr]{1	0.662460747201748\\
2	0.722465385371076\\
3	0.768020284136356\\
4	0.77244925799506\\
5	0.770835874214155\\
};
\addplot [color=mygreen,dashed,style=ultra thick]
  table[row sep=crcr]{1	0.613422410100946\\
2	0.713399017953852\\
3	0.75495128172645\\
4	0.750676580812797\\
5	0.755942067853313\\
};
\addplot [color=mypurple,dotted,style=ultra thick]
  table[row sep=crcr]{1	0.608039659415765\\
2	0.70554145355336\\
3	0.737278821223713\\
4	0.737649156085922\\
5	0.719147069808522\\
};
\addplot [color=myblue,dashdotted,style=ultra thick]
  table[row sep=crcr]{1	0.567232219924697\\
2	0.664562777161069\\
3	0.715976583315309\\
4	0.730565915585498\\
5	0.71330520554789\\
};
\addlegendentry{\topo{*}{3}{32}}
\addlegendentry{\topo{*}{3}{16}}
\addlegendentry{\topo{*}{1}{32}}
\addlegendentry{\topo{*}{1}{16}}
\end{axis}
\end{tikzpicture}%

%% file: figures/LCresults.tex
\begin{table}
\captionsetup{font=scriptsize}
\centering
\caption{
Estimation error for the five sequences $\mathcal{A}$-$\mathcal{E}$ (smaller is better). 
Leaving out the optical map leads to significant deviations (first row) of the error. 
Lane based estimation~\cite{Risack1998} performs better on scenes where the lane is clearly visible ($\mathcal{A}$, $\mathcal{B}$) but has a significant failure in all other conditions ($\mathcal{C}$-$\mathcal{E}$). 
Our approach generates robust estimations for all scenes.
}\label{tab:LCresults}
\begin{tabular}
{c|ccccc}
  Name & \multicolumn{5}{c}{average error [m] short range (0-30\,m)} \\
  & $\mathcal{A}$ & $\mathcal{B}$ & $\mathcal{C}$ & $\mathcal{D}$ & $\mathcal{E}$ \\
  \midrule
  \texttt{w/o optical map} & 1.94 & 2.74 & 3.39 & 2.19 & 2.40 \\
  \midrule
  \topo{1}{1}{16}  & 0.28 & 0.38 & 0.31 & 0.77 & 0.33 \\
  \topo{1}{1}{32}  & 0.27 & 0.33 & 0.27 & 0.64 & 0.29 \\
  \topo{1}{3}{16}  & 0.28 & 0.33 & 0.35 & 0.74 & 0.35 \\
  \topo{1}{3}{32}  & 0.25 & 0.29 & 0.26 & 0.53 & $\mathbf{0.27}$ \\  
  \topo{2}{1}{16}  & 0.25 & 0.30 & 0.35 & 0.73 & 0.31 \\
  \topo{2}{1}{32}  & 0.26 & 0.30 & 0.33 & 0.65 & $\mathbf{0.28}$ \\
  \topo{2}{3}{16}  & 0.26 & 0.31 & 0.33 & 0.64 & 0.31 \\
  \topo{2}{3}{32}  & 0.27 & 0.29 & 0.30 & 0.72 & 0.27 \\
  \topo{3}{1}{16}  & 0.25 & 0.29 & 0.28 & 0.65 & $\mathbf{0.28}$ \\
  \topo{3}{1}{32}  & 0.25 & 0.30 & 0.26 & 0.70 & 0.29 \\
  \topo{3}{3}{16}  & 0.26 & 0.30 & 0.26 & 0.61 & 0.29 \\
  \topo{3}{3}{32}  & 0.26 & 0.31 & 0.29 & 0.64 & 0.28 \\
  \topo{4}{1}{16}  & 0.25 & 0.29 & 0.34 & 0.70 & 0.32 \\
  \topo{4}{1}{32}  & 0.25 & 0.28 & 0.28 & $\mathbf{0.47}$ & 0.29 \\
  \topo{4}{3}{16}  & 0.27 & 0.30 & 0.32 & 0.67 & 0.33 \\
  \topo{4}{3}{32}  & 0.26 & 0.27 & $\mathbf{0.23}$ & 0.57 & $\mathbf{0.27}$ \\
  \topo{5}{1}{16}  & 0.27 & 0.29 & 0.26 & 0.55 & 0.30 \\
  \topo{5}{1}{32}  & 0.26 & 0.29 & 0.31 & 0.61 & $\mathbf{0.28}$ \\
  \topo{5}{3}{16}  & 0.25 & 0.29 & 0.30 & 0.58 & 0.31 \\
  \topo{5}{3}{32}  & 0.26 & 0.28 & 0.28 & 0.54 & $\mathbf{0.28}$ \\
  \midrule
  \texttt{lane-based}~\cite{Risack1998} & $\mathbf{0.19}$ & $\mathbf{0.25}$ & 0.50 & 0.89 & 0.33 \\
  \texttt{failure rate} & 6\,\% & 8\,\% & 65\,\% & 93\,\% & 23\,\% \\
\end{tabular}%
\end{table}

%% file: figures/eyecatcher2.tex
\begin{figure}
\captionsetup{font=scriptsize}
\centering
\includegraphics[width=0.48\textwidth]{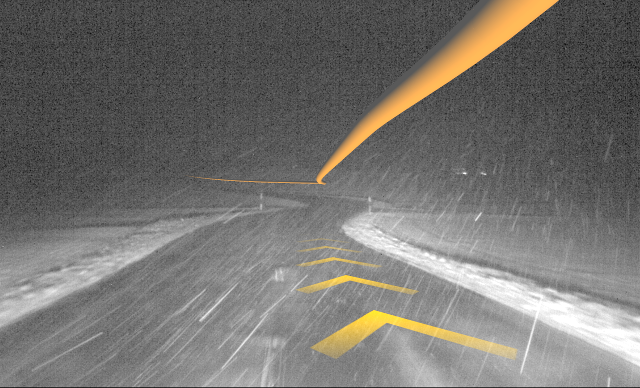}%
\caption{
Augmented reality navigation example for sequence $\mathcal{D}$.
Despite the non-existent lane markings and the heavy image distortions due to weather conditions, the approach proposed in this paper is able to reasonably augment the image.
}\label{fig:eyecatcher2}
\end{figure}

%% file: chapters/summary.tex
\section{Conclusion}\label{sec:conclusion}
This paper presented a deep multi-scale convolutional neural network based approach for camera-based road course prediction and localization.
Various network topologies were trained that reliably detect \road\ and \vehicle\ pixels, from which an optical map is extracted.
Deeper topologies with a higher number of filters per convolution layer perform better, while an increase of pyramid levels after level three does not increase the performance considerably.
The extracted optical maps have been successfully fused with a digital map to refine the lateral localization.
Compared to a baseline lane-based algorithm, the approach proposed in this paper shows a slightly worse performance for optimal road and weather conditions.
However, contrary to the baseline, our approach performs consistently well for various weather conditions, even if lane markings are missing.
This demonstrates that state-of-the-art performance can be achieved while increasing the robustness and application scope to situations, where traditional lane marking detection is not possible.

%% file: chapters/acknowledgments.tex
\section*{ACKNOWLEDGMENT}

The authors would like to thank Markus Thom and Oliver Hartmann for their valuable support.